\documentclass{article}
\usepackage{spconf,amsmath,graphicx}
\usepackage{url}
\usepackage{verbatim}

\let\oldbibliography\thebibliography
\renewcommand{\thebibliography}[1]{%
  \oldbibliography{#1}%
  \setlength{\itemsep}{0pt}%
}

\title{SIMMC: Situated Interactive Multi-Modal Conversational\\Data Collection And Evaluation Platform}
\vspace{-3mm}
\name{Paul A. Crook\sthanks{joint first authors}, Shivani Poddar\footnotemark[\value{footnote}], Ankita De, Semir Shafi\sthanks{work while at Facebook}, David Whitney, Alborz Geramifard, Rajen Subba \vspace{-2mm}}
\address{Facebook Assistant, Facebook AI\\
\small pacrook@fb.com, shivanip@fb.com, deankita@fb.com, semirs@fb.com, dfw@fb.com, alborzg@fb.com, rasubba@fb.com}
\vspace{-11mm}
\begin{document}
%\ninept
%
\vspace{-3mm}
\maketitle
\begin{abstract}
\vspace{-0.5mm}
As digital virtual assistants become ubiquitous, it becomes increasingly important to understand the situated behaviour of users as they interact with these assistants. To this end, we introduce SIMMC, an extension to {ParlAI} for multi-modal conversational data collection and system evaluation. SIMMC simulates an immersive setup, where crowd workers are able to interact with environments constructed in AI Habitat or Unity while engaging in a conversation. The assistant in SIMMC can be a crowd worker or Artificial Intelligent (AI) agent. This enables both \emph{(i)} a multi-player / Wizard of Oz setting for data collection, or \emph{(ii)} a single player mode for model / system evaluation.
We plan to open-source a situated conversational data-set collected on this platform for the Conversational AI research community.

\end{abstract}
\vspace{-2mm}
\section{Introduction}
\vspace{-2mm}
\label{sec:intro}
With digital virtual assistants (VAs) becoming ubiquitous around the home, it is increasingly important to study the situated behaviour of users who are interacting with VAs, \emph{i.e.}\ collect data in more realistic settings, and consider how to train and evaluate VAs that have access to signals beyond unimodal speech or text.
As we outline in Sec.~\ref{sec:relatedwork}, many of the existing large multi-modal (MM) corpora are limited to artificial scenarios, \emph{e.g.}, discussions centered on a single image or video clip, where the user is an observer. These have been seminal in evolving conversational AI from unimodal, to a (now) MM grounding of knowledge. However, it is difficult to translate systems learnt over these corpora to real world scenarios where the user is an actor in the scene. This is a gap that we are aiming to bridge with the use of Virtual Reality (VR) environments. In addition to locational situatedness, we consider that a user's situatedness also pertains to objects in the environment that they are interacting with. To that end a VR setting allows the interactive manipulation of objects.

The primary gaps that we seek to bridge with our work are in making these contributions: \textit{C1. Immersive Data Collection Platform with real time user engagement \& context switching} -- In presenting a Wizard of Oz (Woz) Scenario where both of the annotators are engaged with a VR environment as an assistant and a user, we simulate many of the signals that would be available to a conversational assistant in the real world. Some of the research problems that we target - Multimodal Coreference Resolution, Multimodal Context Tracking, Multimodal Knowledge alignment, Dialog Strategy. \textit{C2. Larger variability in collected scenarios} -- In utilizing VR environments like AI Habitat \cite{habitat19iccv} or Unity \cite{unitygameengine}, we are able to manufacture hundreds of objects for Woz scenarios as compared to real video clips. \textit{C3. Immersive Evaluation platform} -- ParlAI \cite{miller2017parlai} allows pairing of crowd workers for data collection, and a single player mode for evaluation of ML and rule based systems.

\vspace{-3mm}
\section{Related Work}
\vspace{-2mm}
\label{sec:relatedwork}

Language and vision problems such as image captioning, visual question answering
(VQA), and Visual Dialog have gained popularity with the computer vision research
community in recent years, \emph{e.g.}\
% as demonstrated by challenges and associated data sets;
%Conceptual Captions Challenge \cite{}, MSCOCO Image Captioning Challenge \cite{},
VQA challenge/data set \cite{Goyal2019-VQAv2dataset},
Visual Dialog Challenge and VisDial data set \cite{visdial0.9},
\emph{etc.} 
% https://bengio.abracadoudou.com/publications/pdf/vinyals_2016_pami.pdf
% Conceptual Captions: A Cleaned, Hypernymed, Image Alt-text Dataset For Automatic Image Captioning
% Piyush Sharma, Nan Ding, Sebastian Goodman, Radu Soricut
% https://ai.google.com/research/ConceptualCaptions
%
Similarly, in the dialogue community there has been interest in MM dialogues
with DSTC7 and DSTC8 hosting challenge tracks on Audio Video Scene aware dialogues
(AVSD); a data set where two crowd workers have a QA dialog around a short video clip.
In parallel to such work the dialogue research community has a long history of
collecting Woz data sets, %\cite{Kelley1984}
some including MM GUI elements, \emph{e.g.}\ interactive maps \cite{Cheyer1998}.
%\emph{e.g.}\ most recently MULTIWOZ \cite{ramadan2018large}.

%There has been previous exploration of MM Woz \cite{Cheyer1998},
% http://www.adam.cheyer.com/papers/cmc98-1.pdf
%and also evaluation in virtual environments, \emph{e.g.}\ the GIVE challenge \cite{Koller2009}.
% https://www.researchgate.net/publication/220947146_The_Software_Architecture_for_the_First_Challenge_on_Generating_Instructions_in_Virtual_Environments
%SIMMC is similar to both of these. Like Cheyer et al. \cite{Cheyer1998} we proposed a flexible interface that allows the ``user'' and ``wizard'' to share and manipulate a view of the same scene, though in their case the view is a map while ours is rendering of 3D spaces.
%We both provide additional controls to the Wizard to enable them to react to the user's request using whichever modality is the most appropriate.
%Like in the GIVE challenge SIMMC interactions are based in virtual worlds, however we consider a broader range of interactions beyond instruction giving.  

\vspace{-3mm}
\section{Demonstration}
\vspace{-2mm}
\label{sec:main}
In this section, after outlining the overall system architecture, we present two major conversational AI use cases this platform will enable. First, Proactive Assistant Scenarios -- smart proactive recommendations given simulated user context, preferences and background. Second, Interactive Assistant Scenarios -- enabling user interaction with environments in a task oriented MM setting; here furniture shopping.

\begin{figure}[htb]
\vspace{-2mm}
\includegraphics[width=8.7cm, trim=0cm 16.1cm 0cm 0.025cm, clip]{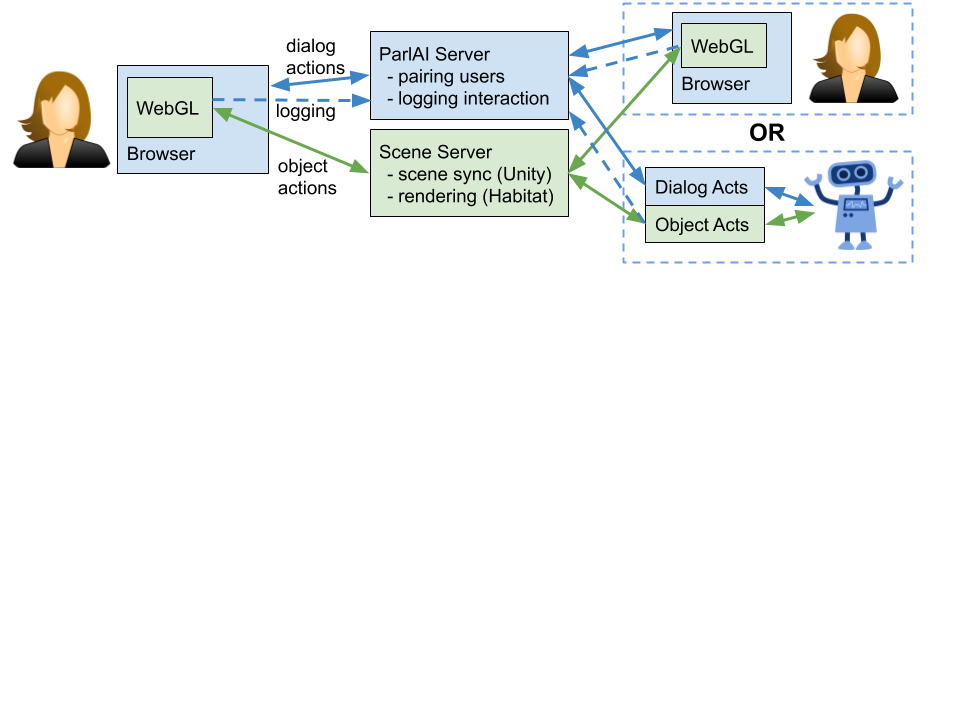}
\vspace{-5mm} 
\caption{SIMMC architecture. Dashed boxes shows alternative arrangements for (i) data collection with a wizard (top), or (ii) evaluation of an AI agent (bottom)}
\label{fig:system}
\vspace{-1mm}
\end{figure}

%\subsection{SIMMC System Architecture}
{\bf SIMMC System Architecture.}
A core of the SIMMC architecture Fig.~\ref{fig:system} is a ParlAI server that in data collection mode matches pairs of crowd workers, or in evaluation mode matches crowd workers and systems.
Each crowd worker sees a message panel where the dialogue is conducted using text or audio and an embedded WebGL view that displays the VR scene. We integrate two VR engines; AI Habitat -- an open source Facebook project to create photorealistic VR scenes, Unity -- a popular game development platform. Both require slightly different integration methods. \emph{AI Habitat} scenes are render remotely by the Scene Server and then frames are streamed to the paired browsers. Key strokes are sent by the WebGL interface to the Scene Server to update the scene. In contrast, \emph{Unity} WebGL instances run locally in each crowd workers' browser. The Unity instances send updates via the Scene Server to ensure paired scenes are kept in sync.
In a Woz data collection setting it is typical to provide additional (GUI) controls to the wizard as part of the scene. This allows them to change the scene in response to the ongoing dialogue with the user.
For both VR engines, events and interactions that occur within the WebGL view are logged back to the ParlAI server. These, along with the dialogue message logs are captured centrally. 
The logs also capture visual object layout information from the Unity/Habitat environments, and screenshots of the scene when each message is exchanged. This provides MM context for user utterances, subsequent assistant responses and actions, and allows model training based on the same.

\begin{figure}[ht]
\begin{minipage}[b]{0.45\linewidth}
\centering
\includegraphics[height=5cm, width=\textwidth]{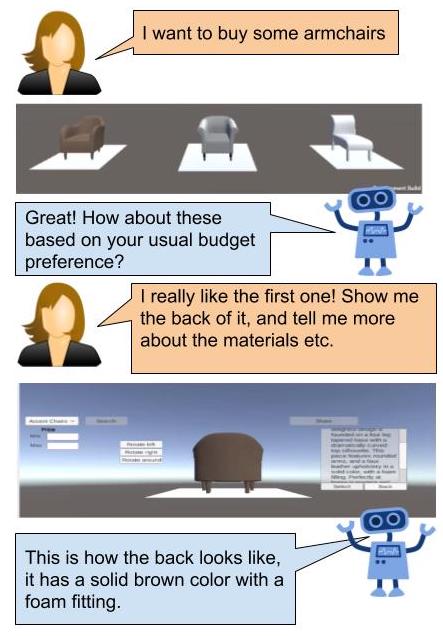}
\caption{Interactive Dialog Example with Unity, for a shopping use case scenario.}
\label{fig:proactive_eg}
\end{minipage}
\hspace{0.5cm}
\begin{minipage}[b]{0.45\linewidth}
\centering
\includegraphics[height=4cm, width=\textwidth]{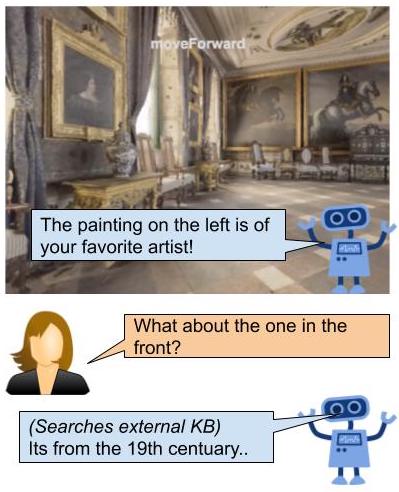}
\caption[Proactive use case example if a user is (say) in a museum, where the assistant can initiate relevant conversation with the user.]{Proactive use case example if a user is (say) in a museum\footnotemark, where the assistant can initiate relevant conversation with the user.}
\label{fig:interactive_eg}
\end{minipage}
\end{figure}

%\subsection{Proactive Situated Data Collection}
{\bf Proactive Situated Data Collection.}
%Data Collection Scenario with AI Habitat
Proactive egocentric scenarios with mapped surroundings is one of the hero use cases that can be collected with SIMMC. Scenarios related to user movement and location in the virtual scenes are enabled by the photo-realistic interiors supported by AI Habitat. Basic scenarios include the assistant providing navigational instructions. By simulating VA knowledge about the user more sophisticated experiences can be enabled such as proactive presentation of information, Fig.~\ref{fig:proactive_eg}, or control of home-automation appliances based on the user's location. 
\begin{comment}
\begin{figure}[htb]
\begin{minipage}[b]{1.0\linewidth}
  \centering
  \centerline{\includegraphics[width=4cm, height=5cm]{pro_hab.jpg}}
\end{minipage}
\caption{Proactive use case example if a user is (say) in a museum, where the assistant can initiate relevant conversation with the user.}
%
\end{figure}
\end{comment}

%\subsection{Interactive Situated Data Collection}
{\bf Interactive Situated Data Collection.}
%Data Collection Scenario with Unity
We can also enable interactive scenarios, wherein a developer can build custom views for the user and assistant to interact with. One such use case would be for (say) a furniture shopping scenario. In the simplest scenario items are presented by the VA based on the user's search criteria, and the user can ask the VA to zoom in and rotate the items, Fig.~\ref{fig:interactive_eg}. The latter allowing us to capture some spacial manipulation of objects by the VA. The user has limited control and has to conduct the majority of the interaction by voice or messages. The Unity game engine allows attributes, such as patterning or colour, of objects to be manipulated increasing the range of conversations that are possible. Although the demonstrated scenario uses furniture one can imagine similar interactions around accessory shopping (\emph{e.g.}\ jewelry, \emph{etc.}).
\footnotetext{CC BY 4.0, Erik Lernest\r{a}l and Sara Dixon,
https://sketchfab.com/3d-models/the-king-s-hall-d18155613363445b9b68c0c67196d98d}
\vspace{-4mm}
\section{Conclusion \& Future Work}
\vspace{-1.5mm}
\label{sec:future}
We demonstrate an immersive, situated user data collection and evaluation platform for conversational AI. This work provides not only the opportunity to collect data sets that move MM conversational modelling towards more realistic scenarios, but we hope it will also inspire additional avenues for conversational AI research, such as proactive conversations.

\bibliographystyle{IEEEbib}
\vspace{-2mm}
\bibliography{refs}

\begin{thebibliography}{1}

\bibitem{habitat19iccv}
{M Savva}, {A Kadian}, {O Maksymets}, Y~Zhao, E~Wijmans, B~Jain, J~Straub,
  J~Liu, V~Koltun, J~Malik, D~Parikh, and D~Batra,
\newblock ``Habitat: {A} {P}latform for {E}mbodied {AI} {R}esearch,''
\newblock in {\em Proc. of ICCV}, 2019.

\bibitem{unitygameengine}
{Unity Technologies},
\newblock ``Unity,'' https://unity.com/, 2019.

\bibitem{miller2017parlai}
A~H {Miller}, W~{Feng}, A~{Fisch}, J~{Lu}, D~{Batra}, A~{Bordes}, D~{Parikh},
  and J~{Weston},
\newblock ``{ParlAI}: A dialog research software platform,''
\newblock {\em arXiv:{1705.06476}}, 2017.

\bibitem{Goyal2019-VQAv2dataset}
Y~Goyal, T~Khot, A~Agrawal, D~Summers-Stay, D~Batra, and D~Parikh,
\newblock ``Making the {V} in {VQA} matter: Elevating the role of image
  understanding in visual question answering,''
\newblock {\em International Journal of Computer Vision}, vol. 127, no. 4, pp.
  398--414, Apr 2019.

\bibitem{visdial0.9}
A~Das, S~Kottur, K~Gupta, A~Singh, D~Yadav, J~Moura, D~Parikh, and D~Batra,
\newblock ``{V}isual {D}ialog,''
\newblock in {\em Proc. of CVPR}, 2017.

\bibitem{Cheyer1998}
A~Cheyer, L~Julia, and J-C Martin,
\newblock ``A unified framework for constructing multimodal experiments and
  applications,''
\newblock in {\em Cooperative Multimodal Communication}, H~Bunt and R-J Beun,
  Eds. 2001, pp. 234--242, Springer Berlin Heidelberg.

\end{thebibliography}

\end{document}